\documentclass{bvm} 
\usepackage{graphicx}
\usepackage{hyperref}

\begin{document}

\newcommand{\bvmyear}{2023}

\selectlanguage{english} 

\title{Comparative Analysis of Radiomic Features and \\Gene Expression Profiles in Histopathology Data \\ Using Graph Neural Networks}


\titlerunning{GNN Analysis of Radiomics and Gene Profiles}

\author{
	Luis Carlos \lname{Rivera Monroy} \inst{1,2}, 
	Leonhard \lname{Rist} \inst{1},
	Martin \lname{Eberhardt} \inst{2}, 
	Christian \lname{Ostalecki} \inst{2}, 
	Andreas \lname{Bauer} \inst{2},
	Julio \lname{Vera} \inst{2},
	Katharina \lname{Breininger} \inst{3}, and
	Andreas \lname{Maier} \inst{1},
}

\authorrunning{Rivera et al.}

\institute{
	\inst{1} Pattern Recognition Lab, Friedrich-Alexander-Universität Erlangen-Nürnberg, Erlangen, Germany\\
	\inst{2} Department of Dermatology, Universitätsklinikum Erlangen, Erlangen, Germany\\
	\inst{3} Department Artificial Intelligence in Biomedical Engineering, FAU Erlangen-Nürnberg, Erlangen, Germany
}

\email{luis.rivera@fau.de}

\maketitle

\begin{abstract}
This study leverages graph neural networks to integrate MELC data with Radiomic-extracted features for melanoma classification, focusing on cell-wise analysis. It assesses the effectiveness of gene expression profiles and Radiomic features, revealing that Radiomic features, particularly when combined with UMAP for dimensionality reduction, significantly enhance classification performance. Notably, using Radiomics contributes to increased diagnostic accuracy and computational efficiency, as it allows for the extraction of critical data from fewer stains, thereby reducing operational costs. This methodology marks an advancement in computational dermatology for melanoma cell classification, setting the stage for future research and potential developments.
\end{abstract}

\section{Introduction}
In post-genomic research, a primary objective is elucidating the relationship between molecular networks and their corresponding cells or tissues. These networks, crucial for specific cellular functions, depend on proteins being precisely positioned, temporally synchronized, and in the correct concentration for effective interactions \cite{Schubert2003}.

Understanding and constructing protein networks by measuring and describing local interactions is vital for deeper understanding of diseases. A significant advancement in this field has been the development of imaging technologies, such as fluoroscopy, which marked a breakthrough by enabling the affinity analysis of multiple proteins. However, the capacity of these technologies to represent the complexity of cellular networks is limited \cite{Gao2021}. Addressing this, methods like Multi-Epitope-Ligand Cartography (MELC) have enhanced the depth of tissue characterization through iterative staining, enabling more comprehensive analysis \cite{Ruetze2010, Bonnekoh2006}.

MELC represents an improvement over traditional imaging techniques by providing detailed tissue characterization at a cellular level. When combined with machine learning, MELC enables detailed cell-wise classification based on gene expression profiles, offering invaluable insights for medical research, including treatment effects and prognosis evolution \cite{Rivera2023, Lazic2020}.

Recent advancements in oncology have leveraged Radiomics combined with machine learning for precision diagnostics. Studies by Liang et al., Pattarone et al., Gomez et al., and Mercaldo et al. highlight the significant role of Radiomics in cancer research, providing new diagnostic and predictive tools through machine-learning classifiers \cite{Liang2023, Pattarone2021, Gomez2022, Mercaldo2022}.

In melanoma, the most lethal skin cancer, prognosis traditionally depends on expert analysis of clinical history and histopathology images \cite{Chopra2020}. Albrecht et al. emphasize the importance of feature standardization in mathematical models for melanoma assessment, considering the subtype diversity and the need for detailed lesion characterization \cite{Albrecht2020}. Graph-based methods are emerging, combining biological parameters with various domain features for a more generalized approach.

This study investigates semi-supervised cell classification in MELC pathology samples by focusing on two types of features: gene expression profiles derived from the multiplex digital image and Radiomics. These features are utilized in both traditional machine learning models and a graph neural network to evaluate their comparative performance. Following the methodology outlined by our previous work \cite{Rivera2023}, this research also incorporates dimensionality reduction techniques to assess their impact on classification accuracy, mainly when applied in conjunction with Graph Neural Networks.

\section{Materials and methods}

\subsection{Dataset}
This investigation evaluated the proposed methodology using a dataset designed for cell-wise classification, comprising tissue samples from suspected melanoma cases. The dataset, included specimens from 27 cases, 20 of which were confirmed melanoma instances, with the remainder being healthy tissue samples \cite{Rivera2023}. Each specimen was processed following the MELC protocol, which involved incubation with affinity reagents, application of fluoroscope-coupled antibodies, capturing images through immunofluorescence microscopy, and gently bleaching the staining agents, allowing for the application of up to 100 different staining agents. In this study, the samples were exposed to 80-85 antigens, and their fluorescent responses were digitized.

The resulting images were digitally captured for analysis, each with a high resolution of 0.45 $\mu m/pixel$ and dimensions of $2018 \times 2018$ pixels. In cases where tumors were identified, expert segmentation was performed by a medical professional, thus maintaining consistency with the approach established in our previous work and facilitating a direct comparison and follow-up of the findings.

\subsection{Graph structure}
In our 2023 study, we demonstrated the effective use of graph-based models to encapsulate complex cellular data, specifically in melanoma \cite{Rivera2023}. This research utilized Cellpose \cite{Pachitariu2022} for cell segmentation, employing the MELC technique to analyze cellular structures. Two distinct graph representations were explored: the first, grounded in feature similarity and spatial proximity, adheres to the methodology established by Wolf et al. \cite{Wolf2018}; the second utilizes cell spatial coordinates, aligning with the approach of Palla et al. \cite{Palla2022}, and adopts a neighborhood size as defined in our previous publication \cite{Rivera2023}.

In these models, individual cells are conceptualized as nodes within a graph, with a typical sample containing approximately 2,080 cells. The relationships between these cells, or edges, are delineated based on the aforementioned criteria of similarity or proximity. For training, the graph is structured to comprise 40,500 nodes and 202,500 edges, providing a comprehensive framework for analyzing cellular interactions in melanoma.

\subsection{Dimensionality Reduction}
Integrating Radiomic features into neighborhood graph-based approaches significantly enhances cell classification in histopathological analysis. These features include first-order intensity statistics, shape-based descriptors, and complex texture features. Notably, texture features from the Gray Level Co-occurrence Matrix (GLCM) and the Gray Level Run Length Matrix (GLRLM) provide a detailed characterization of cellular attributes \cite{Griethuysen2017}. GLCM analyzes the spatial relationship of pixel intensities, while GLRLM focuses on the length of consecutive pixels having the same intensity in an image. When applied as node attributes in graphs, these features impart a nuanced understanding of each cell, addressing challenges like initialization sensitivity and suboptimal local minima in graph optimization \cite{Ardizzoni2023}.

Furthermore, advanced dimensionality reduction techniques such as uniform manifold approximation and projection (UMAP) and t-distributed stochastic neighbor edging (tSNE) enhance this process. These techniques effectively compress high-dimensional data, facilitating the creation of detailed yet computationally efficient neighborhood graphs \cite{Wang2021, Do2021}. This combination of image analysis and graph-based modeling improves cell classification accuracy and provides deeper insights into biological structures in histopathological images \cite{Pati2020}. The resulting graph models, enriched with Radiomic data, become potent tools for identifying patterns in cell populations, enhancing diagnostic precision and research in histopathology \cite{Wang2021, Do2021}.

\subsection{Graph Random Neural Network}
In the domain of semi-supervised learning for graph-structured data, the Graph Random Neural Networks (GRAND) architecture, as introduced by Feng et al. (2020), has demonstrated notable efficacy, particularly in weakly supervised tasks, robust performance, and efficient training \cite{Feng2020}. This architecture addresses critical limitations inherent in traditional graph neural networks (GNNs), such as over-smoothing, non-robustness, and inadequate generalization in scenarios with sparse labeled nodes. GRAND distinguishes itself by strategically decoupling feature propagation from transformation, a design choice that mitigates complexity and over-smoothing issues prevalent in conventional GNNs. Central to its success in graph-structured data analysis, GRAND's methodology integrates a random propagation strategy using DropNode to create perturbed feature matrices, enhancing robustness and employing consistency regularization to maintain uniform classification confidence across nodes and their neighborhoods.

\begin{figure}[h!]    
	\centering 
	\includegraphics[width=\textwidth]{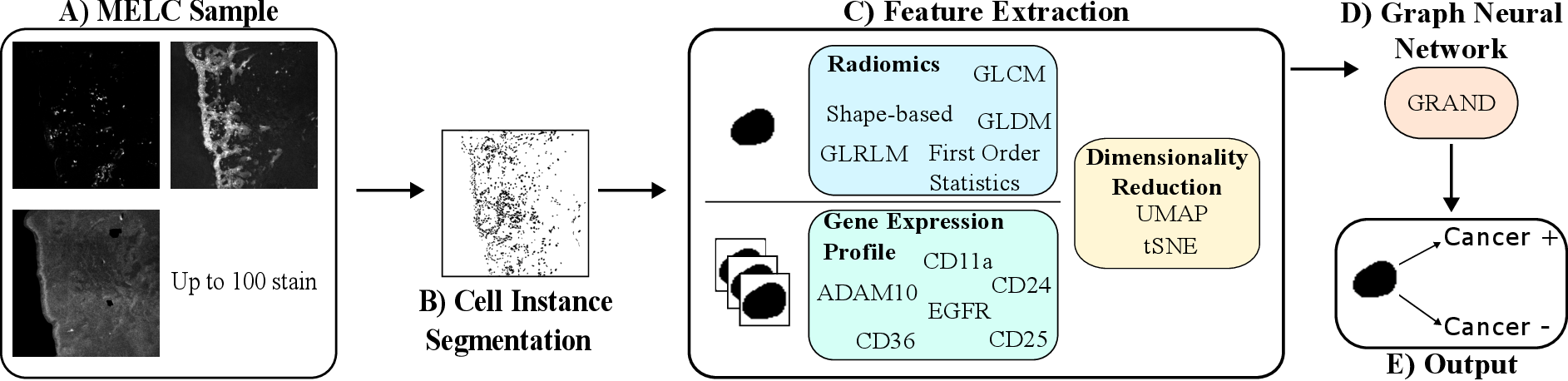}	
    \caption{Computational Pipeline for MELC Data-Based Cell Classification. A) Input: Multiple MELC sample images. B) Semi-supervised cell instance segmentation. C) Feature Extraction: Radiomic features and Gene expression profiling. D) Features encoded in a Graph Neural Network (GRAND). E) Output: Classification of cells as cancer positive or negative}
    \label{fig:pipeline}
\end{figure}

\subsection{Experimental Design}
In this study, we adhered to the methodology for performance evaluation as outlined by our previous work, employing a data split of 70\% for training, 10\% for validation, and 20\% for testing \cite{Rivera2023}. This distribution was designed to ensure a balanced representation of healthy and melanoma tissues across each dataset segment. The optimization of hyperparameters was methodically conducted on the training set, utilizing a Bayesian search strategy.

We employed XGBoost and Random Forest for baseline comparisons, widely recognized as standard machine-learning algorithms for handling tabular data. The graph neural network (GNN) training followed the original authors' implementation guidelines and a similar approach to hyperparameter optimization \cite{Feng2020}. Our analysis aims to compare the efficacy of two distinct types of features: gene expression profiles derived from multi-array imaging and Radiomic-generated features, which encompass first-order intensity statistics, shape-based descriptors, and complex texture features. This comparative study is designed to elucidate these feature sets' relative strengths and applications in the context of our machine learning models. An overview of the proposed pipeline can be seen in Figure \ref{fig:pipeline}.
\section{Results}
As outlined in Table \ref{tab:results}, a discernible trend indicates enhanced performance in node classification for melanoma cells when utilizing Radiomic features combined with UMAP for training. This observation is drawn from a comparative evaluation of two feature types in the GNN framework and their interactions with various dimensionality reduction techniques.

\section{Discussion}
The analysis of our results, as detailed in Table \ref{tab:results}, demonstrates that graph representations incorporating cell spatial locations outperform those based on gene expression profiles alone. This finding supports the importance of spatial context in cell classification. We observed that methods using tabular profiles and feature graphs show similar performance, attributed to their comparable information encoding strategies.

In dimensionality reduction, UMAP emerges as the superior technique, aligning with previous findings on its ability to capture spatial patterns and cellular neighborhoods effectively \cite{Rivera2023}. Conversely, tSNE focuses on maintaining feature correlations, offering value but less impacting classification performance.

A pivotal discovery in our study is the consistent effectiveness of Radiomic features over direct gene expression profiles from digital pathological samples. This trend is consistent across various methods and types of dimensionality reduction, and Radiomic features achieve enhanced results when combined with these techniques. Addressing limitations in staining agent selection, Radiomic-extracted features emerge as a viable solution, extracting relevant biological features for melanoma cell classification. This research marks an advancement in computational dermatology, suggesting new avenues for more precise and efficient melanoma cell classification techniques.

\begin{table}[h!]    
	\centering 
	\caption{ \textbf{Comparative Analysis of Gene Expression Profiles and Radiomic Features in Training Graph Convolutional Neural Networks.} This table summarizes the results of training graph convolutional neural networks using two distinct data sources: gene expression profiles derived from images and features extracted via Radiomics.}
	\includegraphics[scale=0.465]{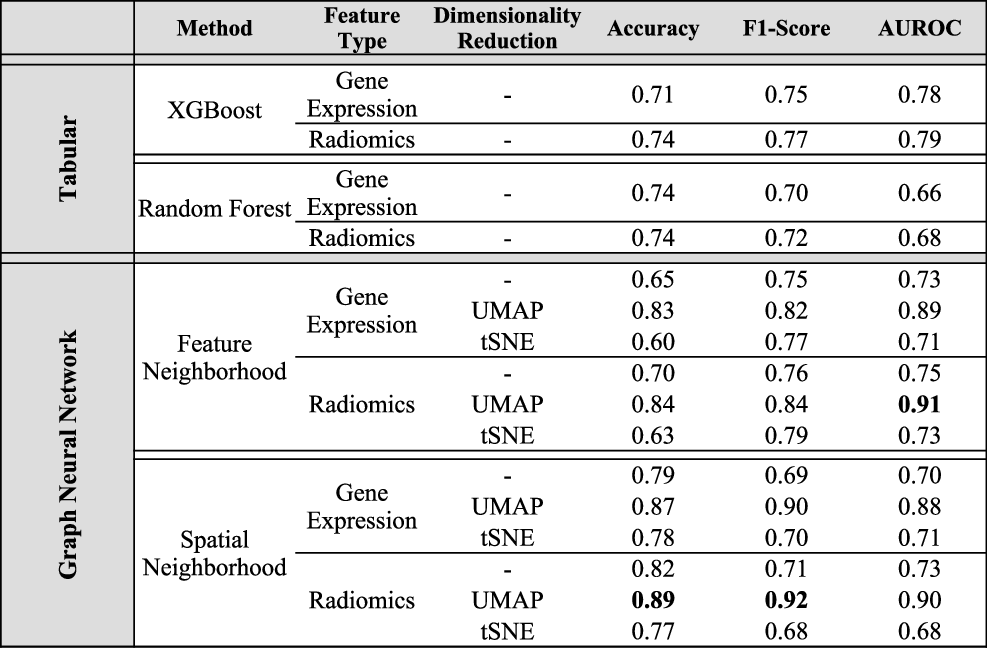}	
    \label{tab:results}
\end{table}

\printbibliography

@inbook{Schubert2003,
	author    = {Schubert, Walter},
	title     = {Topological Proteomics, Toponomics, MELK-Technology},
	booktitle = {Proteomics of Microorganisms: Fundamental Aspects and Application},
	year      = {2003},
	publisher = {Springer Berlin Heidelberg},
	address   = {Berlin, Heidelberg},
	pages     = {189--209},
	isbn      = {978-3-540-36459-7},
	doi       = {10.1007/3-540-36459-5_8},
	url       = {https://doi.org/10.1007/3-540-36459-5_8}
}

@article{Gao2021,
  title={Quantitative Fluorescence Resonance Energy Transfer Analysis on the Direct Interaction of Activation-2b with Histone H3/Switch-3B Protein in Arabidopsis Mesophyll Protoplasts},
  author={Gao, Lu and Lin, Fangrui and Han, Danlu and Jiang, Jieming and Yang, Chengwei and Zhuang, Zhengfei and Chen, Tongsheng},
  journal={J Fluoresc},
  pages={981--988},
  year={2021},
  publisher={Springer}
}

@article{Ruetze2010,
	author = {Ruetze, Martin and Gallinat, Stefan and Wenck, Horst and Deppert, Wolfgang and Knott, Anja},
	title = {In situ localization of epidermal stem cells using a novel multi epitope ligand cartography approach},
	journal = {Integrative Biology},
	volume = {2},
	number = {5-6},
	pages = {241-249},
	year = {2010},
	month = {04},
	issn = {1757-9708},
	doi = {10.1039/b926147h},
	url = {https://doi.org/10.1039/b926147h},
	eprint = {https://academic.oup.com/ib/article-pdf/2/5-6/241/27327821/b926147h.pdf}
}

@article{Bonnekoh2006,
	author = {Bonnekoh, B. and Böckelmann, R. and Pommer, A.J. and Malykh, Y. and Philipsen, L. and Gollnick, H.},
	title = {The CD11a Binding Site of Efalizumab in Psoriatic Skin Tissue as Analyzed by Multi-Epitope Ligand Cartography Robot Technology: Introduction of a Novel Biological Drug-Binding Biochip Assay},
	journal = {Skin Pharmacol Physiol},
	volume = {20},
	number = {2},
	pages = {96-111},
	year = {2006},
	month = {12},
	issn = {1660-5527},
	doi = {10.1159/000097982},
	url = {https://doi.org/10.1159/000097982},
	eprint = {https://karger.com/spp/article-pdf/20/2/96/3544223/000097982.pdf}
}

@INPROCEEDINGS{Rivera2023,
  author={Rivera Monroy, Luis Carlos and Rist, Leonhard and Eberhardt, Martin and Ostalecki, Christian and Baur, Andreas and Vera, Julio and Breininger, Katharina and Maier, Andreas},
  booktitle={2023 IEEE 20th International Symposium on Biomedical Imaging (ISBI)}, 
  title={Employing Graph Representations for Cell-Level Characterization of Melanoma MELC Samples}, 
  year={2023},
  volume={},
  number={},
  pages={1-5},
  doi={10.1109/ISBI53787.2023.10230519}
}

@article{Lazic2020,
  title={Single-cell landscape of bone marrow metastases in human neuroblastoma unraveled by deep multiplex imaging},
  author={Lazic, Daria and Kromp, Florian and Kirr, Michael and Mivalt, Filip and Rifatbegovic, Fikret and Halbritter, Florian and Bernkopf, Marie and Bileck, Andrea and Ussowicz, Marek and Ambros, Inge M and others},
  journal={bioRxiv},
  pages={2020--09},
  year={2020},
  publisher={Cold Spring Harbor Laboratory}
}

@article{Liang2023,
  author = {Liang, Wenhua and Wang, Bo and Tao, Jinsheng and Peng, Minhua and Tu, Xixiang and Qiu, Xiangcheng and Yang, Yang and Ye, Zhujia and Chen, Zhiwei and Fan, Jianbing and He, Jianxing},
  title = {{A machine learning–based multidimensional model integrating clinical, radiomics, and cell-free DNA methylation biomarkers for the classification of pulmonary nodules.}},
  journal = {Journal of Clinical Oncology},
  volume = {41},
  number = {16\_suppl},
  pages = {3070-3070},
  year = {2023},
  doi = {10.1200/JCO.2023.41.16\_suppl.3070} 
}

@article{Pattarone2021,
  title={Learning deep features for dead and living breast cancer cell classification without staining},
  author={Pattarone, Gisela and Acion, Laura and Simian, Marina and Mertelsmann, Roland and Follo, Marie and Iarussi, Emmanuel},
  journal={Sci Rep},
  volume={11},
  number={1},
  pages={10304},
  year={2021},
  publisher={Nature Publishing Group UK London}
}

@article{Gomez2022,
  title={Analysis of cross-combinations of feature selection and machine-learning classification methods based on [18F] F-FDG PET/CT radiomic features for metabolic response prediction of metastatic breast cancer lesions},
  author={G{\'o}mez, Ober Van and Herraiz, Joaquin L and Ud{\'\i}as, Jos{\'e} Manuel and Haug, Alexander and Papp, Laszlo and Cioni, Dania and Neri, Emanuele},
  journal={Cancers},
  volume={14},
  number={12},
  pages={2922},
  year={2022},
  publisher={MDPI}
}

@article{Mercaldo2022,
  title={Prostate Gleason Score Detection by Calibrated Machine Learning Classification through Radiomic Features},
  author={Mercaldo, Francesco and Brunese, Maria Chiara and Merolla, Francesco and Rocca, Aldo and Zappia, Marcello and Santone, Antonella},
  journal={Applied Sciences},
  volume={12},
  number={23},
  pages={11900},
  year={2022},
  publisher={MDPI}
}

@article{Chopra2020,
  title={Pathology of melanoma},
  author={Chopra, Asmita and Sharma, Rohit and Rao, Uma NM},
  journal={Surgical Clinics},
  volume={100},
  number={1},
  pages={43--59},
  year={2020},
  publisher={Elsevier}
}

@article{Albrecht2020,
  title={Computational models of melanoma},
  author={Albrecht, Marco and Lucarelli, Philippe and Kulms, Dagmar and Sauter, Thomas},
  journal={Theoretical Biology and Medical Modelling},
  volume={17},
  number={1},
  pages={1--16},
  year={2020},
  publisher={BioMed Central}
}

@article{Feng2020,
  title={Graph random neural networks for semi-supervised learning on graphs},
  author={Feng, Wenzheng and Zhang, Jie and Dong, Yuxiao and Han, Yu and Luan, Huanbo and Xu, Qian and Yang, Qiang and Kharlamov, Evgeny and Tang, Jie},
  journal={Adv Neural Inf Process Syst},
  volume={33},
  pages={22092--22103},
  year={2020}
}

@article{Wang2021,
  title={SUSCC: secondary construction of feature space based on UMAP for rapid and accurate clustering large-scale single cell RNA-seq data},
  author={Wang, Hai-Yun and Zhao, Jian-ping and Zheng, Chun-Hou},
  journal={Interdisciplinary Sciences: Computational Life Sciences},
  volume={13},
  pages={83--90},
  year={2021},
  publisher={Springer}
}

@article{Do2021,
  title={A generalization of t-SNE and UMAP to single-cell multimodal omics},
  author={Do, Van Hoan and Canzar, Stefan},
  journal={Genome Biol},
  volume={22},
  number={1},
  pages={1--9},
  year={2021},
  publisher={BioMed Central}
}

@article{Pachitariu2022,
  title={Cellpose 2.0: how to train your own model},
  author={Pachitariu, Marius and Stringer, Carsen},
  journal={Nat Methods},
  volume={19},
  number={12},
  pages={1634--1641},
  year={2022},
  publisher={Nature Publishing Group US New York}
}

@article{Wolf2018,
  title={SCANPY: large-scale single-cell gene expression data analysis},
  author={Wolf, F Alexander and Angerer, Philipp and Theis, Fabian J},
  journal={Genome Biol},
  volume={19},
  pages={1--5},
  year={2018},
  publisher={Springer}
}

@article{Palla2022,
  title={Squidpy: a scalable framework for spatial omics analysis},
  author={Palla, Giovanni and Spitzer, Hannah and Klein, Michal and Fischer, David and Schaar, Anna Christina and Kuemmerle, Louis Benedikt and Rybakov, Sergei and Ibarra, Ignacio L and Holmberg, Olle and Virshup, Isaac and others},
  journal={Nat Methods},
  volume={19},
  number={2},
  pages={171--178},
  year={2022},
  publisher={Nature Publishing Group US New York}
}

@misc{Ardizzoni2023,
      title={Local Optimization of MAPF solutions on Directed Graphs}, 
      author={S. Ardizzoni and I. Saccani and L. Consolini and M. Locatelli},
      year={2023},
      eprint={2304.01765},
      archivePrefix={arXiv},
      primaryClass={math.OC}
}

@article{Griethuysen2017,
    author = {van Griethuysen, Joost J.M. and Fedorov, Andriy and Parmar, Chintan and Hosny, Ahmed and Aucoin, Nicole and Narayan, Vivek and Beets-Tan, Regina G.H. and Fillion-Robin, Jean-Christophe and Pieper, Steve and Aerts, Hugo J.W.L.},
    title = "{Computational Radiomics System to Decode the Radiographic Phenotype}",
    journal = {Cancer Res},
    volume = {77},
    number = {21},
    pages = {e104-e107},
    year = {2017},
    month = {10},
    issn = {0008-5472},
    doi = {10.1158/0008-5472.CAN-17-0339},
    url = {https://doi.org/10.1158/0008-5472.CAN-17-0339},
    eprint = {https://aacrjournals.org/cancerres/article-pdf/77/21/e104/2934659/e104.pdf},
}

@inproceedings{Pati2020,
  title={Hact-net: A hierarchical cell-to-tissue graph neural network for histopathological image classification},
  author={Pati, Pushpak and Jaume, Guillaume and Fernandes, Lauren Alisha and Foncubierta-Rodr{\'\i}guez, Antonio and Feroce, Florinda and Anniciello, Anna Maria and Scognamiglio, Giosue and Brancati, Nadia and Riccio, Daniel and Di Bonito, Maurizio and others},
  booktitle={Uncertainty for Safe Utilization of Machine Learning in Medical Imaging, and Graphs in Biomedical Image Analysis},
  pages={208--219},
  year={2020},
  organization={Springer}
}

\end{document}